\documentclass[11pt]{article}
\usepackage{emnlp2016}
\usepackage{times}
\usepackage{url}
\usepackage{latexsym}
\usepackage{xspace}
\usepackage{booktabs}
\usepackage{color}
\usepackage{graphicx}
\usepackage{tabulary}
\usepackage{enumitem}
\usepackage{amsfonts}

\newenvironment{tightitemize}%
  {\begin{itemize}[topsep=0pt, partopsep=0pt] %
    \setlength{\itemsep}{0pt}%
    \setlength{\parskip}{0pt}%
    }%
  {\end{itemize}}
  
  \usepackage{color}
  {\begin{enumerate}≈ \footnotesize%
    \setlength{\itemsep}{0pt}%
    \setlength{\parskip}{0pt}%
    }%
  {\end{enumerate}}

\usepackage{microtype}
\usepackage{multirow}
\usepackage{verbatim}
\usepackage{amsmath,amsthm,amssymb}
\usepackage{array}
\usepackage[scaled=0.86]{helvet}
\usepackage{ifthen}
\usepackage{courier}
\usepackage[linesnumbered,vlined,ruled]{algorithm2e}

\newcommand{\captionfonts}{\small}
\makeatletter  
\long\def\@makecaption#1#2{%
  \vskip\abovecaptionskip
  \sbox\@tempboxa{{\captionfonts #1: #2}}%
  \ifdim \wd\@tempboxa >\hsize
    {\captionfonts #1: #2\par}
  \else
    \hbox to\hsize{\hfil\box\@tempboxa\hfil}%
  \fi
  \vskip\belowcaptionskip}
\makeatother   

\belowdisplayskip \abovedisplayskip

\DeclareMathOperator*{\argmax}{arg\,max}

\emnlpfinalcopy

\newcommand{\eos}{{\it EOS}\xspace}
\newcommand{\sts}{{{\textsc{Seq2Seq}}}\xspace}

\title{A Simple, Fast Diverse Decoding Algorithm for Neural Generation}

\author{Jiwei Li, Will Monroe and Dan Jurafsky\\
Computer Science Department, Stanford University, Stanford, CA, USA \\
{\tt jiweil,wmonroe4,jurafsky@stanford.edu} 
}
\date{}

\begin{document}
\maketitle

\begin{abstract}
We propose a simple, fast  decoding algorithm that fosters diversity in neural generation. 
The algorithm modifies the standard beam search algorithm by penalizing
hypotheses that are siblings---expansions of the
same parent node in the search---thus
favoring including hypotheses from diverse parents. 
We evaluate the model on three neural generation tasks:
dialogue response generation,
abstractive summarization,
and machine translation. 
We also describe an extended model that
uses reinforcement learning to
automatically choose the appropriate level of beam diversity
for different inputs or tasks.
Simple diverse decoding helps across all three tasks, 
especially those needing reranking or having diverse
ground truth outputs; reinforcement learning offers an additional boost.  
\footnote{This paper  includes material from the unpublished manuscript  ``Mutual Information and Diverse Decoding Improve Neural Machine Translation'' (Li and Jurafsky, 2016).}
\end{abstract}
\section{Introduction}
Neural generation models \cite{sutskever2014sequence,bahdanau2014neural,cho2014learning,kalchbrenner2013recurrent}
are of growing interest for various applications such as machine translation 
\cite{sennrich2015neural,gulcehre2015using}, conversational response generation 
\cite{vinyals2015neural,sordoni2015neural,luan2016lstm}, 
abstractive summarization \cite{nallapatiabstractive,rush2015neural,chopra2016abstractive}, and image caption generation
\cite{chen2015microsoft}.
Such models are trained by learning to predict an output sequence,
and then
at test time, the model chooses the best sequence given the input,
usually using  beam search.

One long-recognized issue with beam search is lack of diversity in the beam:
candidates often differ only by punctuation
or minor morphological variations, with most of the words overlapping
\cite{macherey2008lattice,tromble2008lattice,kumar2004minimum}.
Lack of diversity can hinder sequence generation quality.
For tasks like conversational response generation or image caption generation,
there is no one correct answer;
the decoder thus needs to explore different paths to various sequences to avoid 
local minima \cite{vijayakumar2016diverse}.

Lack of diversity causes particular problems in two-stage
re-ranking approaches,  in which
an N-best list or lattice of candidates is generated using beam search 
and then re-ranked using features too global or
expensive to include in the first beam decoding pass
\footnote{E.g., position bias or bilingual attention symmetry in MT  \cite{cohn2016incorporating} or   global  discourse features in summarization.}.
In neural response generation, a re-ranking step helps avoid generating dull or
generic responses \cite{li2015diversity,sordoni2015neural,shao15}.
Lack of diversity in the N-best list significantly decreases the impact of
reranking.\footnote{\newcite{shao15}, for example, find that
re-ranking heuristics  in conversational response generation 
work for shorter responses but not for long responses, since the  beam N-best list
for long responses are mostly identical even with a large beam.}

\begin{figure*}
\includegraphics[width=2in]{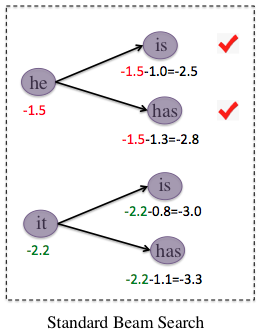}
\includegraphics[width=2.6in]{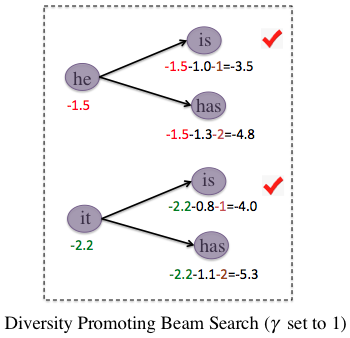}
\centering
\caption{An illustration of standard beam search and the proposed diversity-promoting beam search. $\gamma$ denotes the hyperparameter for penalizing intra-sibling ranking. Scores are made up for illustration purposes. }
\label{figure}
\end{figure*}

In this paper, we propose a simple, fast, diversity-fostering beam search model for neural decoding;
the model can be obtained by changing just one line of beam search code in MATLAB.
The algorithm uses standard beam search as its backbone but adds an additional
term penalizing siblings---expansions of the same parent node in the search---
thus favoring choosing hypotheses from diverse parents (as demonstrated in Figure \ref{figure}).

The proposed model supports batched decoding using GPUs, significantly speeding up the decoding process compared to other diversity fostering models for phrase-based MT systems 
\cite{macherey2008lattice,tromble2008lattice,kumar2004minimum,devlin2012trait}. 
To show the generality of the model, we evaluate  it on three 
neural generation tasks---conversational response generation, abstractive summarization and 
machine translation--
demonstrating that the algorithm generates better outputs due to considering more diverse sequences.
We discuss which properties of these various tasks make them more or less likely
to be helped by diverse decoding.
We also propose a more sophisticated variant that uses reinforcement learning
to automatically adjust the diversity rate for different inputs, yielding an additional performance boost.



\section {Related Work}
Diverse decoding has been sufficiently explored in phrase-based MT
\cite{huang2008forest,finkel2006solving}, including
use of compact representations like lattices and hypergraphs
\cite{macherey2008lattice,tromble2008lattice,kumar2004minimum},
``traits'' like translation length 
\cite{devlin2012trait}, bagging/boosting
\cite{xiao2013bagging}, blending multiple systems \cite{cer2013positive},
and sampling translations proportional
to their probability \cite{chatterjee2010minimum}.
The most relevant is work from  
\newcite{gimpel2013systematic} and \newcite{batra2012diverse}   that
produces diverse N-best lists by adding a dissimilarity function
based on N-gram overlaps, distancing the current translation from already-generated ones
by choosing translations that have high scores but are distinct from previous ones.
While we draw on these intuitions, these
existing diversity-promoting algorithms are tailored to phrase-based translation frameworks and not easily transplanted to neural MT decoding, which requires batched computation.

Some recent work has looked at decoding for neural generation. 
\newcite{cho2016noisy}
proposed 
 a meta-algorithm that
runs in parallel many chains of the noisy version of an inner decoding algorithm.
\newcite{vijayakumar2016diverse} 
 proposed a diversity-augmented objective
for image caption generation
 akin to a neural version of \newcite{gimpel2013systematic}.
\newcite{shao15}  
used a stochastic search algorithm 
that reranks the hypothesis segment by segment, which injects diversity earlier in the decoding process.

The proposed RL based algorithm is inspired by a variety of recent reinforcement learning approaches in NLP for tasks such as dialogue \cite{dhingra2016end}, word compositions \cite{dani16}, machine translation \cite{ranzato2015sequence}, neural model visualization \cite{lei2016rationalizing}, and coreference \cite{clark2016deep}.

\section{Diverse Beam Decoding}
In this section, we introduce the proposed algorithm. We first go over the vanilla beam search method and then detail the proposed algorithm which fosters diversity during decoding.
\subsection{Basics}
Let $X$ denote the source input, which is the input dialogue history for conversational response generation or a
source sentence for machine translation.  The input $X$ 
is mapped to a vector representation, which is used as the initial input to the decoder.
Each $X$ is paired with a target sequence $Y$, which corresponds to a dialogue utterance in response generation
or
 a target sentence in machine translation. 
$Y=\{y_1,y_2,...,y_{n_y}\}$ 
 consists a sequence of $n_y$ words.
A neural generation model 
 defines a distribution over outputs and sequentially predicts tokens using a softmax function:
\begin{equation*}
\begin{aligned}
p(Y|X)
&=\prod_{t=1}^{n_y}p(y_t|X,y_1,y_2,...,y_{t-1})\\
\end{aligned}
\label{equ-lstm}
\end{equation*}
At test time, the goal is to find the sequence $Y^*$ that maximizes the probability given input $X$:
\begin{equation}
Y^*=\argmax_{X}p(Y^*|X)
\end{equation}

\subsection{Standard Beam Search for N-best lists} 

N-best lists  are 
standardly
generated from a model of $p(Y|X)$ using a beam search decoder.
As illustrated in Figure 1, at time step $t-1$ in decoding, the decoder keeps track of $K$ hypotheses, where $K$ denotes the beam size, and their scores $S(Y_{t-1}|X)=\log p(y_1,y_2,...,y_{t-1}|X)$. As it moves on to time step $t$,
it expands each of the $K$ hypotheses
(denoted as $Y_{t-1}^k=\{y_1^k,y_2^k,...,y_{t-1}^k\}$, $k\in [1,K]$)
 by selecting the top $K$ candidate expansions,  each expansion
 denoted as  $y_t^{k,k'}$, $k'\in [1,K]$,
  leading to the construction of $K\times K$ new hypotheses:
  $$[Y_{t-1}^k, y_t^{k,k'}], k\in [1,K], k'\in [1,K]$$
The score for each of the $K\times K$ hypotheses is computed as follows:
\begin{equation}
S(Y_{t-1}^k,y_t^{k,k'}|x)=S(Y_{t-1}^k|x)+\log p(y_t^{k,k'}|x,Y_{t-1}^k)
\end{equation}
In a standard beam search model, the top $K$ 
hypotheses are selected
(from the $K\times K$  hypotheses computed in the last step) based on 
the score $S(Y_{t-1}^k,y_t^{k,k'}|x)$. The remaining hypotheses are ignored when the algorithm proceeds to the next time step.

\subsection{Generating a Diverse N-best List}
Unfortunately, the N-best lists outputted from standard beam search are a poor surrogate 
for the entire search space \cite{finkel2006solving,huang2008forest}. 
The beam search algorithm can only keep a small proportion of candidates in the search space, and
most of the generated translations in N-best list are similar.
Our proposal is to increase diversity by changing
the way $S(Y_{t-1}^k,y_t^{k,k'}|x)$ is computed,
as shown in Figure \ref{figure}.
For each of the  hypotheses $Y_{t-1}^k$ ({\it he} and {\it it}),
we generate the top $K$ translations
 $y_t^{k,k'}$, $k'\in [1,K]$ as in the standard beam search model.
Next, we  rank the $K$ translated tokens generated from the same parental hypothesis
based on $p(y_t^{k,k'}|x,Y_{t-1}^k)$ 
in descending order: {\it he is} ranks first among {\it he is} and {\it he has}, and {\it he has} ranks second;
similarly for {\it it is} and {\it it has}.
 
We then rewrite the score for $[Y_{t-1}^k, y_t^{k,k'}]$ by adding an additional term $\gamma k'$,
 where $k'$ denotes the ranking of the current hypothesis among its siblings (1 for {\it he is} and {\it it is}, 2 for {\it he has} and {\it it has}).
\begin{equation}
\hat{S}(Y_{t-1}^k,y_t^{k,k'}|x)=S(Y_{t-1}^k,y_t^{k,k'}|x)-\gamma k'
\label{dive}
\end{equation}
We call $\gamma$ the {\it diversity rate}; it indicates the degree of diversity one wants to integrate into the beam search model.

The top $K$ hypotheses are selected based on $\hat{S}(Y_{t-1}^k,y_t^{k,k'}|x)$ as we move on to the next time step.
By adding the additional term $\gamma k'$, 
the model punishes lower-ranked hypotheses among siblings (hypotheses descended from the same parent).
When we compare newly generated hypotheses descended from different ancestors, the model gives more credit to  top hypotheses from each of the different ancestors.
For instance, even though the original score for {\it it is} is lower than {\it he has},
the model favors the former as the latter is more severely punished by the intra-sibling ranking part $\gamma k'$. 
The model thus generally favors choosing hypotheses from diverse parents, leading to a more diverse N-best list. 
The proposed model is straightforwardly implemented with a minor adjustment to the standard beam search.
\section{Automatically Learning Diversity Rate}
One disadvantage of the algorithm described above is that a fixed diversity rate $\gamma$ is applied to all examples.   Yet the optimal  diversity rate could vary from instance to instance,
and too high a diversity could even be detrimental if it pushes the decoding model 
too far from the beam search scores. Indeed,
\newcite{shao15} find that in response generation, 
standard beam search works well for short responses but deteriorates as the sequence gets longer,
while
\newcite{vijayakumar2016diverse} argue in image caption generation that diverse decoding is 
beneficial for images with many objects, but not images with few objects.

A good diverse decoding algorithm should have the ability to automatically adjust its diversity rates for different inputs---for example, using small diversity rates for images with fewer objects but larger rates for those with more objects. 
We propose a reinforcement learning-based algorithm called
{\it diverseRL} that is capable of learning different $\gamma$ values for different inputs. 
\subsection{Model}
We first define a list $\Gamma$ that contains the values that $\gamma$ can take. 
For example, $\Gamma$ might consist of the 21 values in the range [0,1] at regularly spaced intervals 0.05 apart.\footnote{This is just for illustration purpose. One can define any set of diversity rate values.}
Our main idea is to use reinforcement learning (policy gradient methods) to discover the best diversity rate  $\gamma(X)$ for a given input $X$ with respect to
the final evaluation metric. 
For each input $X$, we parameterize the action of choosing an associated diversity rate $\gamma(X)$ by a policy network $\pi(\gamma(X)=\gamma'|X)$
which is a distribution over the $|\Gamma|$ classes.\footnote{
This means the probability of $\gamma(X)$ taking on two similar values (e.g., 0.05 and 0.1) are independent.  
An alternative is to make $\gamma$ continuous. However, we find that
using discrete values is good enough because of the large amount of training data, and discrete values are easier to implement.}
We first map the input $X$ to a vector representation $h_X$ using a recurrent net\footnote{This recurrent net  shares  parameters with the standard generation model.}, and then map $h_X$ to a policy distribution over different values of $\gamma$ using a softmax function: 
\begin{equation}
\pi(\gamma(X)=\gamma'|X)=\frac{\exp( h_X^T\cdot h_{\gamma'})}{\sum_{j=1}^{j=|\Gamma|}\exp( h_X^T\cdot h_{\Gamma_j})}
\label{softmax}
\end{equation}
Given an action, namely a choice of $\gamma'$ for $\gamma(X)$, we start decoding using the proposed diverse decoding algorithm and obtain an N-best list.
Then we pick the best output---the output with the largest reranking score, or the output with the largest probability if no reranking is needed.
Using the selected output, we compute the evaluation score (e.g., BLEU)
denoted $R(\gamma(X)=\gamma')$, and this score is used as the reward\footnote{This idea is inspired by recent work  \cite{ranzato2015sequence} that uses BLEU score as reward in reinforcement learning for machine translation. Our focus is different  since we are only interested in learning the policy to obtain diversity  rates $\gamma(X)$.}
 for the action of choosing diversity rate $\gamma(X)=\gamma'$. 

We use the REINFORCE algorithm \cite{williams1992simple}, a kind of policy gradient method, to find the optimal diversity rate policy by maximizing the expectation of the final reward, denoted as follows:
\begin{equation}
E_{\pi(\gamma(X)=\gamma'|X)} [ R(\gamma(X)=\gamma')) ]
\end{equation}
The expectation is approximated by sampling from $\pi$ and the gradient is computed based on 
 the likelihood ratio \cite{glynn1987likelilood,aleksand1968stochastic}:
 \begin{equation}
\begin{aligned}
\nabla E(\theta)=[R(\gamma(X)-b] \nabla\log \pi(\gamma(X)=\gamma'|X)
\end{aligned}
\end{equation}
where $b$ denotes the baseline value.\footnote{The baseline value is estimated using another neural model that takes as input $X$ and outputs a scalar $b$ denoting the estimation of the reward.
The baseline model is trained by minimizing the mean squared loss between the estimated reward $b$
and actual cumulative reward $r$, $||r-b||^2$.
We refer the readers to \cite{ranzato2015sequence,zaremba2015reinforcement} for more details.
The baseline estimator model is independent from the policy models and the
error is not backpropagated back to them.
}
The model is trained to take actions (choosing a diversity rate $\gamma$) that will lead to the highest value of final rewards. 
\subsection{Training and Testing}

To learn the policy $\pi(\gamma(X)=\gamma'))$, we take a pretrained encoder-decoder model and run  additional epochs over the training set in which we keep the encoder-decoder parameters fixed and do the diverse beam search using $\gamma(X)$ sampled from the policy distribution. Because decoding is needed for every training sample, training is extremely time-consuming. Luckily, since the sentence composition model
involved in 
$\pi(\gamma(X)=\gamma|X)$
shares parameters with the pre-trained encoder-decoder model, the only parameters needed to be learned are those in the softmax function in Eq.~\ref{softmax},  the number of which  is relatively small.
We therefore only take a small fraction of training examples (around 100,000 instances).

Special attention is needed for tasks in which feature-based reranking is used for picking the final output:
 feature weights will change as $\gamma$ changes because those weights are tuned based on the decoded N-best list for dev set,
   usually using MERT \cite{och2003minimum}. Different $\gamma$ will lead to different dev set N-best lists and consequently different feature weights.
We thus adjust feature weights using the dev set after every 10,000 instances. Training takes roughly 1 day. 

\section{Experiments}
We conduct experiments on three different sequence generation tasks: conversational response generation, abstractive summarization and machine translation, the details of which are described  below. 
\subsection{Conversational Response Generation}

\subsubsection{Dataset and Training}
We used 
the OpenSubtitles 
(OSDb) dataset \cite{tiedemann2009news}, an  open-domain movie script dataset containing roughly 60M-70M scripted lines spoken by movie characters. Our models
are trained to predict the current turn given the preceding
ones. 
We trained a two-layer encoder-decoder  model with attention \cite{bahdanau2014neural,luong2015effective},  with 512 units in each layer. 
We treat the two preceding dialogue utterances as dialogue history,
simply concatenating them to form the source input. 

To better illustrate in which scenarios the proposed  algorithm offers the most help, 
we construct three 
different 
dev-test splits
 based on reference length, specifically:
\begin{tightitemize}
\item  {\it Natural}: Instances randomly sampled from the dataset.
\item {\it Short}: Instances with target reference length no greater than 6.
\item {\it Long}: Instances with target reference length greater than 16.
\end{tightitemize}
Each set  contains roughly 2,000 instances. 

\subsubsection{Decoding and Reranking}
\noindent 
We consider the following two settings:
\paragraph{Non-Reranking} No reranking is needed. We simply pick the output with highest probability using standard and diverse  beam search. 
\paragraph{Reranking} 
Following \newcite{li2015diversity}, we first
generate an N-best list using vanilla or diverse beam search and 
 rerank the generated responses
by  combining  likelihood $\log p(Y|X)$, backward likelihood $\log p(X|Y)$,\footnote{$\log p(Y|X)$ is trained in a similar way as standard \sts models with only sources and targets being swapped. 
} sequence length $L(Y)$, and language model likelihood $p(Y)$. 
The linear combination of $\log (Y|X)$ and $\log p(X|Y)$ is a generalization of the mutual information between the source and the target, which
dramatically decreases the rate
 of dull and generic responses.
.
Feature weights are optimized using 
MERT \cite{och2003minimum} on N-best lists of response
candidates.\footnote{We set the minimum length and  maximum length
of a decoded target
 to 0.75 and 1.5 times the length of sources. 
Beam size K is set to 10.
We then rerank the N-best list and pick the output target with largest final ranking score. 
}

\subsubsection{Evaluation}
For automatic evaluations, 
we report (1) BLEU \cite{papineni2002bleu},
 which has widely been used in response generation evaluation; 
and (2) diversity,
which is
the number of distinct unigrams and bigrams in
generated responses scaled by the total
number of generated tokens.
This scaling is 
 to avoid favoring long
sentences, as described in Li et al. (2016a).

\begin{table}
\centering
\small
\begin{tabular}{cccc}
&Natural&Short&Long \\\hline
\multicolumn{4} {c} {Without-Reranking} \\\hline
vanilla& 1.30 &1.59 &0.89 \\
diverse&1.42&1.63&1.06 \\
&  (+$9.1\%$)  &  (+$2.5\%$) &  (+$19\%$) \\
diverseRL & 1.49& 1.67& 1.03    \\
&(+$15\%$)   & (+$5.0\%$)&    (+$16\%$)\\\hline
\multicolumn{4} {c} {With-Reranking} \\\hline
vanilla& 1.88&2.52&1.17   \\
diverse& 2.21 &2.75&1.68 \\  
& (+$17\%$) &(+$9.2\%$)&(+$44\%$) \\
diverseRL &2.32&2.79&1.70     \\\
& (+$23\%$) &(+$11\%$)&(+$47\%$) \\\hline
\end{tabular}
\caption{Response generation: BLEU scores from the vanilla beam search model, the proposed diversity-promoting beam search model 
and the diverse Beam search + Reinforcement Learning (diverseRL) model
on various datasets.}
\label{BLEU}
\end{table}
 
 \begin{table}
\centering
\small
\begin{tabular}{cccc}
&Natural&Short&Long \\\hline
\multicolumn{4} {c} {Without-Reranking} \\\hline
Vanilla& 0.032  & 0.079 &0.009 \\
 Diverse&0.049   & 0.101& 0.049 \\
&(+$53.2\%$) & (+$27.8\%$) &   (+$445\%$)  \\\hline
 \multicolumn{4} {c} {With-Reranking} \\\hline
Vanilla& 0.068   &0.252 &0.025  \\
Diverse &0.095 &0.282 &0.070 \\
& (+$39.7\%$)& (+$12.2\%$) &  (+$180\%$)  \\\hline
\end{tabular}
\caption{Response generation: Diversity scores from standard beam search model and the proposed diversity-promoting beam search model on various datasets.
\label{diversity}}
\end{table}

 \begin{table}
\centering
\small
\begin{tabular}{cccc}
Diverse-Win&Diverse-Lose&Tie \\\hline
62\%&16\%&22\% \\
\end{tabular}
\caption{Response generation: The gains over the standard beam search model from diverseRL
based on pairwise human judgments.}
\label{human}
\end{table}

 We also do human evaluation, as suggested by \newcite{liu2016not}.
We employ crowdsourced
judges to provide evaluations for a random
sample of 200 items. Each output pair was ranked
by 3 judges, who were asked to decide which of the
two outputs was better. They were instructed to prefer
outputs that were more specific (relevant) to the context. Ties were permitted. Identical
strings were automatically assigned the same
score. 
 
 \begin{table}[t]
 \centering
\scriptsize
 \begin{tabular}{ll}\hline
 {\bf Input}& {\bf As the joint chief of the eight clans, you should do something.}\\\hline
 {\bf N-best List} & If you want to fight, you must go to the temple.    \\
({\bf Diverse Beam}) &If you want to fight , you can do it yourself . \\
& No, i'm not going to do anything. \\
 & I'm not going to let you go, i'm not going to let you do that. \\
& But i don 't want to be a monk \\\hline
 {\bf N-best List} &  If you want to fight, you must go to the temple . . . . \\
 ({\bf Vanilla Beam}) & if you want to fight , you must go to the temple to fight .   \\
 & If you want to fight , you must go to the temple . . . \\
 & If you want to fight , you must go to the police station . \\
 & If you want to fight , you must go to the palace . \\\hline\hline
 {\bf Input}& {\bf  Well , why can 't you just believe us? }\\\hline 
  {\bf N-best List} & 
    I mean, it 's not like we 're going to be together forever . \\
  ({\bf Diverse Beam}) & 
    I mean, you 're the one who 's been stealing my stuff . 
 \\
  & Because i 'm not a good person . \\
&   
   Why can 't we just go to the police and get the hell out of here ? \\
& This is a very important moment for us .  \\\hline 
 {\bf N-best List}  & I mean , it 's not like we 're going to be together forever .     \\
  ({\bf Vanilla Beam}) & I mean , it 's not like we 're gonna be together forever .   \\
  & I mean , it 's not like we 're going to be together any more . \\
&  I mean , it 's not like we 're going to be together any more , is it ? \\
 & I mean , it 's not like we 're going to be friends forever . \\\hline\hline
  {\bf Input}& {\bf  Senator , why didn 't you come to me with this information ?}\\\hline
   {\bf N-best List} &  I didn 't want you to think i was crazy .\\
     ({\bf Diverse Beam}) &  We could have been killed . \\
&     Because you 're not a threat to me . \\
& You know , i was thinking about the way you handled the situation . \\
& If you 're not gonna tell me, i'm not gonna tell you .\\\hline 
 {\bf N-best List}  &  I didn 't want you to think i was crazy .     \\
  ({\bf Vanilla Beam}) &  I didn 't want to upset you .  \\
  & I didn 't want you to worry . \\
  &I didn 't want you to find out. \\
  & I didn 't want you to worry about me . \\\hline
  \end{tabular}
  \caption{Response generation: Sample responses using the diversity-promoting beam search and vanilla beam search.}
  \label{sample-response}
 \end{table}

\subsubsection{Results}
Results for BLEU scores and diversity scores are presented in Tables \ref{BLEU} and \ref{diversity}.  
Diverse decoding boosts performance across all settings, but more so in the {\it reranking} settings than the {\it non-reranking} ones. 
For {\it reranking}, the improvement is much larger for longer responses than shorter ones.   We suspect that this is because short responses have a smaller
search space, and so the vanilla beam search  has a sufficiently diverse beam.
For longer responses, the generator may get trapped in a 
local decoding minimum, and as \cite{shao15} pointed out can
generate incoherent or even contradictory responses like {\it ``I like fish but I don't like fish"} .
Table \ref{sample-response} shows that
for these longer responses, these hypotheses generated by standard beam search are nearly identical,
dramatically decreasing the impact of the reranking process.
\footnote{Sampling could be another way of generating diverse responses. However,  responses from sampling are usually incoherent, especially for long sequences. This is because the error accumulates as decoding goes on. A similar phenomenon has also been observed by \newcite{shao15}. }

On the {\it Natural} set, which has the same distribution of length as the full data, we see a performance boost 
from the {\it diverseRL} model's ability to dynamically adjust diversity.
For the {\it short} and {\it long} sets,
the improvement from the diverseRL model is less significant, 
presumably because the
 dataset has been pre-processed in such a  way that examples are more similar to each other in terms of length. 

In terms of human evaluation, we find that the {\it diverseRL} model  produces responses with better general quality, winning 62 percent of the time when compared to standard beam search.  

\subsection{Abstractive Summarization}
\subsubsection{Training and Dataset}
We consider two settings for abstractive summarization. For the first setting (denoted {\it single}), we follow the protocols described in \newcite{rush2015neural}, in which  the source input is the first sentence of the document
to summarize. We train a word-level  attention  model for this setting. 
Our training dataset consists of 800K pairs. 

A good summarization system should have the ability 
to   summarize a large chunk of text, separating
wheat from chaff. We thus  consider another setting in which each input consists of multiple sentences (denoted {\it multi}).\footnote{
We consider 10 sentences at most for each input.
Sentence position treated as a word feature which is associated with an embedding to be learned as suggested in \newcite{nallapatiabstractive}}
We train a hierarchical model with attention at sentence level \cite{li2015hierarchical} for generation, which has been shown to yield better results than word-level encoder-decoder models for multi-sentence summarization \cite{nallapatiabstractive}.

 For reranking,
we employ various  global features taken from or  inspired by 
 prior non-neural work in summarization \cite{daume2006bayesian,nenkova2005impact,mckeown1999towards}.
 Features we consider include (1) average tf-idf score of constituent words in the generated output; (2) {\it KLSum} \cite{haghighi2009exploring}, which reflects the topic 
 distribution 
 overlap between the entire input document and the generated output;\footnote{
Specifically, the KL divergence between the topic distributions assigned by a variant of the LDA model \cite{blei2003latent} that identifies general, document-specific and topic-specific word clusters.} and
   (3) backward probability $p(X|Y)$, i.e., the probability of generating the entire document given the summary. 
   For evaluation,
 we report ROUGE-2 scores  \cite{lin2004rouge} in Table \ref{ROUGE}.
 
\begin{table}
\centering
\small
\begin{tabular}{ccc}
&Single &Multi \\\hline
\multicolumn{3} {c} {Without-Reranking} \\\hline
vanilla& 11.2 &8.1  \\
diverse&12.1&9.0 \\
\multicolumn{3} {c} {With-Reranking} \\\hline
vanilla& 12.4&9.3   \\
diverse& 14.0 &11.5 \\  \hline
\end{tabular}
\caption{ROUGE-2 scores from vanilla beam search model and the proposed diversity-promoting beam search model for abstractive summarization.}
\label{ROUGE}
\end{table}

The proposed diverse reranking algorithm helps in both cases, but more significantly in the {\it reranking} setting than the {\it non-reranking} one. 
This is because the mechanisms by which diverse decoding helps 
in the two settings
 are different:
 for the {\it non-reranking} setting, if the conditional entropy in the target distribution is  low enough, the decoded string from standard beam search will be close to the global optimum, and thus there  won't be much space for improvement. 
For  the {\it reranking} setting, however, the story is different:
it is not just about finding the global optimal output based on the target distribution, but also about
 incorporating different criteria to make up for facets that are missed by the encoder-decoder model. This requires the N-best list to be diverse for the reranking models to make a significant difference. 

We also find an improvement from the {\it reranking} model using document-level features over the {\it non-reranking} model.  
We did not observe a big performance boost from the {\it diverseRL} model over the standard diverse model (around 0.3 ROUGE score boost) on this task. This is probably because
the {\it diverseRL} model helps most when inputs are different and thus need different diverse decoding rates. 
In this task,
 the input documents are all news articles and  share similar properties, so a unified diversity rate tuned on the dev set might already be good enough.  
Results for {\it diverseRL} are thus omitted for brevity.  

Interestingly, we find that the result for the {\it multi} setting is significantly worse than the {\it single} setting, which is also observed by \newcite{nallapatiabstractive}: adding more sentences leads to worse results. This illustrates the incapability of neural generation models to date to summarize long documents. The proposed diverse decoding model produces a huge performance boost in the {\it multi} setting. 

\subsection{Machine Translation}
\subsubsection{Dataset and Training}
The models are trained on the WMT'14 training dataset containing 4.5 million pairs for English-German.
We limit our vocabulary to
the top 50K most frequent words for both languages. 
Words not in the vocabulary are replaced by a universal unknown token. 
We use newstest2013 (3000 sentence pairs) as the development set and report translation performances in BLEU  \cite{papineni2002bleu} on newstest2014 (2737 sentences). 
We  trained neural \sts models \cite{sutskever2014sequence} with  attention \cite{luong2015effective,cho2014learning}.
Unknown words are replaced
using methods similar to those of  \newcite{luong2015effective}. 
\subsubsection{Decoding and Reranking}
Again we consider both  {\it reranking} settings and {\it non-reranking} settings, 
where  for  {\it non-reranking} we  select the best output using standard beam search. 
For {\it reranking}, 
we first generate a large N-best list using the beam search algorithm.\footnote{We use beam size $K=50$ both for standard beam search and diverse beam search. At each time step of decoding, we are presented with $K\times K$ word candidates. 
We first add all hypotheses with an \eos token generated at the current time step to the N-best list. 
Next, we preserve the top K unfinished hypotheses and move to the next time step. 
We therefore maintain constant batch size as hypotheses are completed and removed, by adding in more unfinished 
hypotheses. This allows the size of final N-best list for each input to be much larger than the beam size.}
We then rerank the hypotheses using features
that have been shown to be useful in neural machine translation \cite{sennrich2015improving,gulcehre2015using,cohn2016incorporating,cheng2015agreement}, including (1) backward probability $p(X|Y)$; (2) language model probability $p(Y)$ trained from monolingual data \cite{gulcehre2015using,gulcehre2015using};\footnote{
$p(t)$ is trained using a single-layer LSTM recurrent models  
 using monolingual data. 
 We use News Crawl corpora from WMT13 (\url{http://www.statmt.org/wmt13/translation-task.html}) as additional training
data to train monolingual language models. 
We used a subset of the original dataset which 
 roughly contains 60 millions sentences. 
} (3) bilingual symmetry: the agreement between attentions in German-English and English-German; and
(4) target length. Again feature weights are optimized using 
MERT \cite{och2003minimum} on N-best lists of response
candidates in the dev set.

\begin{table}
\centering
\small
\begin{tabular}{ll}\hline
\multicolumn{2} {c} {Non-Reranking} \\
Vanilla beam  & 19.8   \\
Diverse & 19.8 (+0.04)  \\ 
Diverse+RL & 20.2 (+0.25)  \\ \hline
\multicolumn{2} {c} {Reranking}\\
Vanilla   & 21.5 \\
Diverse  &22.1 (+0.6) \\
Diverse+RL  &22.4 (+0.9) \\\hline
\end{tabular}
\caption{BLEU scores from vanilla beam search model and the proposed diversity-promoting beam search
 model  on WMT'14 English-German translation.}
 \label{BLEU-MT}
\end{table}

\subsubsection{Results}
Experimental results are shown in Table \ref{BLEU-MT}. 
First, 
no significant  improvement is observed for the proposed diverse beam search model in the {\it non-reranking} setting. 
The explanation is similar to the  abstractive summarization task:
the vanilla beam search algorithm with large beam size is already good enough at finding 
the global optimum, and the small 
benefit from diverse decoding for some examples might even be canceled out by others where diversity rate value is too large. The {\it diverseRL} model addresses the second issue, leading to a  performance boost of +0.25. 
For the {\it reranking} setting, the performance boost is more significant, +0.6 for diverse beam search and +0.9 for {\it DiverseRL}. 

Overall, diverse decoding doesn't seem to improve machine translation 
as much as it does summarization and response generation. 
We suspect this is due to the very low 
entropy of the  target distribution (perplexity less than 6), so
standard beam search is already fairly strong.\footnote{The difference between beam size 2 and beam size 12 in decoding for French-English translation is only around 0.3 BLEU score \cite{sutskever2014sequence}, which also confirms that the standard decoding algorithm is sufficiently powerful for MT.}
\section{Discussion}
In this paper, we introduce a general diversity-promoting decoding algorithm for neural generation. The model adds an intra-sibling ranking term to the standard beam search algorithm, favoring choosing hypotheses from diverse parents. 
 The proposed model is a general, simple  and fast algorithm 
 that will bring a performance boost to all neural generation tasks for which a diverse N-best list is needed.

On top of this approach, we build a more sophisticated algorithm 
that is capable of 
 automatically adjusting diversity rates for different inputs using reinforcement learning. We find that, at the expense of model complexity and training time (compared with the basic RL-free diverse decoding algorithm), the model is able to adjust its diversity rate to better values, yielding generation quality better than either standard beam search or the basic diverse decoding approach.

Diverse decoding doesn't help all tasks equally,
contributing the most in two kinds of tasks:
(1) Those with a very diverse space of ground truth outputs 
(e.g., tasks like response generation), rather than those
in which the conditional entropy of the target distribution is already low enough (e.g., machine translation)
(2) Tasks for which reranking is needed to
incorporate features not included in the first decoder pass,
such as document-level  abstractive summarization.

~~\\
\noindent {\bf {\large Acknowledgements}}

\begin{small}

We would especially like to 
thank Thang Luong Minh for insightful discussions and releasing relevant code,
as well as Sida Wang, Ziang Xie, Chris Manning and other members of the Stanford NLP group for helpful comments and suggestions. 
We also thank Michel Galley, Bill Dolan, Chris Brockett, Jianfeng Gao 
and other members of the NLP group at Microsoft Research 
for helpful discussions.  Jiwei Li is supported by a Facebook Fellowship,  which we gratefully acknowledge. 
This work is partially supported by the DARPA Communicating
with Computers (CwC) program under ARO prime
contract no. W911NF- 15-1-0462 and the NSF via  IIS-1514268.
Any opinions,
findings, and conclusions or recommendations expressed
in this material are those of the authors and
do not necessarily reflect the views of DARPA, the NSF,
or Facebook.

\end{small}

\bibliographystyle{acl2012}
\bibliography{MMI_MT}

\end{document}